\documentclass[a4paper]{article}
\usepackage{Odyssey2020}
\usepackage{epsfig,amssymb,amsmath}
\usepackage{times}
\usepackage{graphicx}
\usepackage{multirow}
\usepackage{latexsym}
\usepackage{booktabs}

\newcommand\Model{\textsc{BERTphone}}

\def\recon{\lambda}
\newcommand\TaskSys{x-vector + SAP model}

\usepackage[hyperfootnotes=false]{hyperref}
\usepackage{cleveref}
\usepackage{amsfonts}
\usepackage{xcolor}
\usepackage[stable]{footmisc}

\title{{\Model}: Phonetically-aware Encoder Representations for Utterance-level Speaker and Language Recognition}

\name{Shaoshi Ling, Julian Salazar, Yuzong Liu, Katrin Kirchhoff}

\address{Amazon AWS AI \\
{\small \tt \{shaosl, julsal, liuyuzon, katrinki\}@amazon.com} }

\begin{document}
\maketitle

\begin{abstract}
We introduce {\Model}, a Transformer encoder trained on large speech corpora that outputs phonetically-aware contextual representation vectors that can be used for both speaker and language recognition. This is accomplished by training on two objectives: the first, inspired by adapting BERT to the continuous domain, involves masking spans of input frames and reconstructing the whole sequence for acoustic representation learning; the second, inspired by the success of bottleneck features from ASR, is a sequence-level CTC loss applied to phoneme labels for phonetic representation learning. We pretrain two {\Model} models (one on Fisher and one on TED-LIUM) and use them as feature extractors into x-vector-style DNNs for both tasks. We attain a state-of-the-art $C_{\text{avg}}$ of 6.16 on the challenging LRE07 3sec closed-set language recognition task. On Fisher and VoxCeleb speaker recognition tasks, we see an 18\% relative reduction in speaker EER when training on {\Model} vectors instead of MFCCs. In general, {\Model} outperforms previous phonetic pretraining approaches on the same data. We release our code and models at \url{https://github.com/awslabs/speech-representations}.
\end{abstract}

\section{Introduction}
\label{sec:introduction}

\begin{figure*}[t]
  \centering
  \includegraphics[trim=0 650 0 0,clip,width=0.7\linewidth]{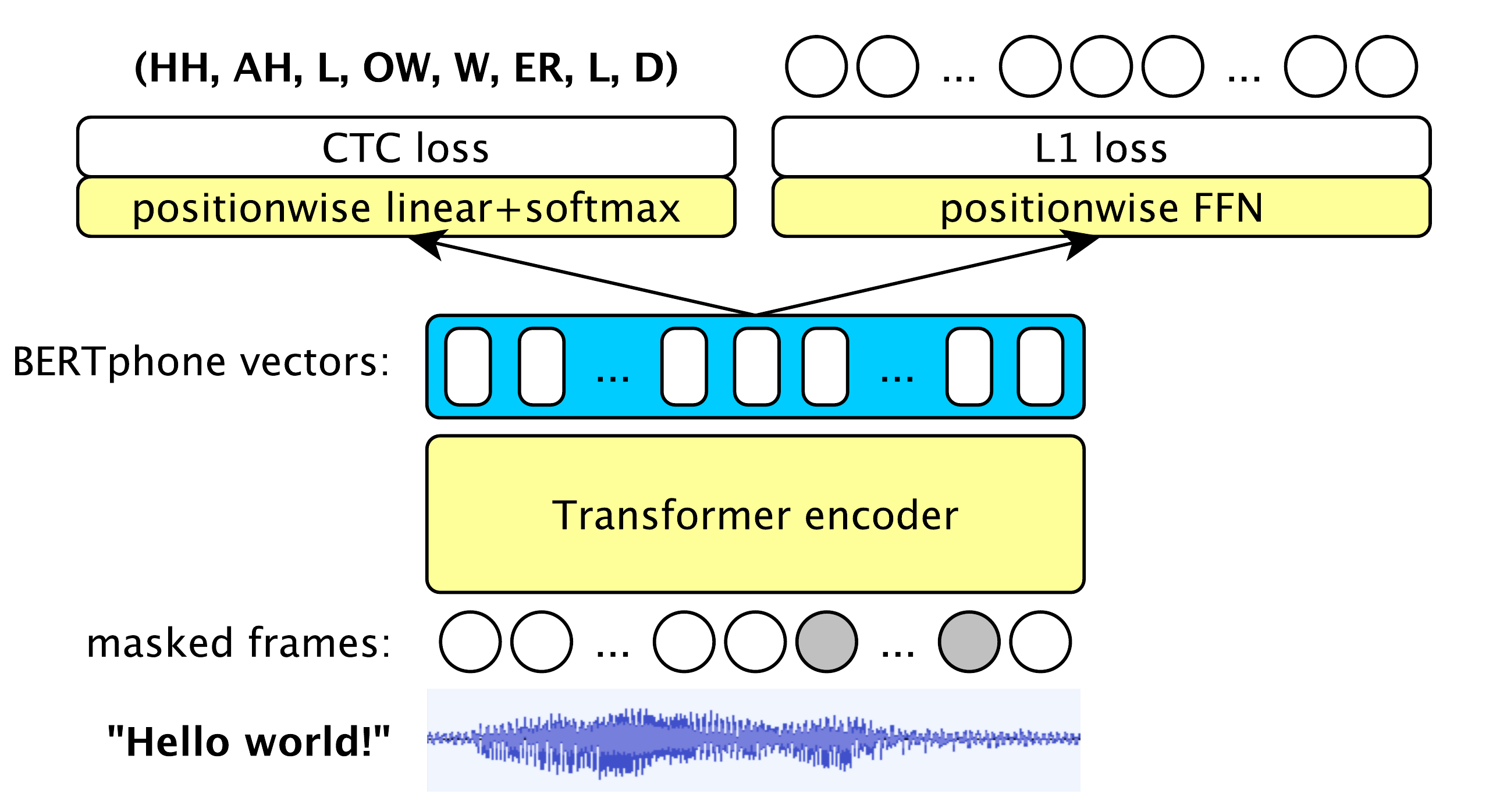}
  \caption{{\Model} pretraining. Contiguous spans of features are masked to zero, as represented by the gray circles. Two objectives are applied: one reconstructs the original frames from before masking, while the other predicts context-independent phoneme labels for each frame then computes CTC loss against the ground truth phoneme sequence.}
  \label{fig:model_view}
\end{figure*}

Motivated by improvements from pretraining bidirectional contextual representations like BERT \cite{devlin2018bert} for language understanding tasks, we propose {\Model}, a pretraining approach that gives versatile representations for speech processing tasks. It is a deep Transformer encoder \cite{vaswani2017attention} that turns an utterance into a frame-wise vector sequence capturing both acoustic and phonetic information. Notably, {\Model} can be used as a fixed feature extractor on which task-specific models for speaker, language, and speech recognition can be trained.

BERT is a deep neural network trained on text under a masked language modeling (MLM) objective, where subword tokens are masked then predicted via classification. In contrast, speech features are continuous, which suggests their reconstruction as a possible generative analogue. However, acoustic frames are highly correlated over time, which models may exploit during training \cite{chung2019unsupervised}. More generally, a contrastive loss may be better at inducing high-level latent knowledge \cite{oord2018cpc}. For example, phonetic understanding in spoken language recognition is often incorporated via intermediate ``bottleneck'' representations from neural networks trained to classify context-dependent phonemes \cite{matejka2014bottleneck}.

Hence, {\Model} uses two objectives to pretrain on large amounts of speech data, which we describe in\Cref{sec:Architecture}. First, to induce higher-level acoustic representations, we propose masking contiguous \textit{spans} of acoustic frames then reconstructing all frames independently under an $L_1$ penalty. Second, to induce and implicitly align phonetic information to each frame, we propose applying connectionist temporal classification (CTC) \cite{graves2006ctc}, a sequence loss computed against each utterance's unaligned context-independent phoneme labels.

After pretraining, we use the final layer of {\Model} as a fixed featurizer for a self-attentional x-vector model (\Cref{sec:task}) that is trained for either (text-independent) speaker recognition (SR) \cite{snyder2018spoken} or language recognition (LR) \cite{snyder2018spoken1}. In \Cref{sec:experiments,sec:results}, we pretrain two {\Model} models (Fisher, TED-LIUM) then train classifiers for each of SR and LR; altogether, we consistently improve on end-to-end systems and other phonetic pre-training methods. Finally, in \Cref{sec:analysis} we explore different weightings of both losses, train over {\Model's} layers, and try graphemes as an alternate supervision for CTC. {\Model} gives competitive phoneme error rates, suggesting downstream use for end-to-end automatic speech recognition (ASR) as well.

\section{Related work}

\subsection{Phonetic information for SR and LR}

Over the past decade, the speech processing community has adopted deep neural networks (DNNs) for ASR, in both hybrid (HMM-DNN) \cite{hinton2012dnn} and end-to-end configurations \cite{graves2006ctc, chan2016las}. Past works have explored the ``indirect method'' of using the hybrid system's DNN as a frame-level feature extractor for SR and LR \cite{richardson2015dnn}. As the DNN is already trained on large speech corpora, it is thought to have captured higher-level phonetic understanding that is useful for LR \cite{house1977lr}, and to a lesser extent for SR \cite{lei2014novel, wang2019usage}. This is often done with \textit{bottleneck features} \cite{matejka2014bottleneck} (more generally, intermediate frame-wise features from a DNN), though DNN posteriors have also been used as sufficient statistics for i-vector systems \cite{lei2014novel}. However, since SR can require higher-level acoustic information (speaker traits) that are less relevant for ASR, the DNN's knowledge may be insufficient. This has led to original speech features being presented in \textit{tandem} with DNN features in SR systems \cite{richardson2015dnn}. Another approach is to multitask the frame-level DNN with SR directly \cite{liu2018speaker}.

In contrast, we train an end-to-end ASR system as our feature extractor (removing the need for forced alignment) to capture phonetic knowledge. We only use the final layer for both tasks. To enable this, instead of multitasking with SR or presenting early layers in tandem, we use a self-supervised reconstruction task to induce the desired acoustic knowledge.

\subsection{Deep acoustic representations} 

Recent works have improved the effectiveness of DNNs in ASR by pretraining on large amounts of unlabeled speech. These have involved recurrent neural networks with self-supervised objectives that are either \textit{contrastive} (classifying a future audio sample from negative examples) or \textit{reconstructive} (recreating future audio frames or wave samples). A contrastive loss was used by CPC \cite{oord2018cpc} for phone classification and SR, and wav2vec \cite{schneider2019wav2vec} for ASR. Full reconstruction was used for autoencoding by \cite{chung2016audio}, and future frame reconstruction was used in autoregressive predictive coding (APC) \cite{chung2019unsupervised} for phone classification and SR.

Concurrent to our work, a number of preprints have also proposed pretraining speech representations bidirectionally (with recurrent networks or Transformer encoders), with either a contrastive \cite{kawakami2019unsupervised, baevski2019vq} or reconstructive \cite{song2019speech, liu2019mockingjay, 1912.01679} objective. None of these explore the benefits of weak or multi-task supervision, which we use to enable frozen, multi-purpose representations. These works also do not evaluate LR or large-scale out-of-corpus SR (VoxCeleb), which are the focus of our work.

\section{{\Model} pretraining}
\label{sec:Architecture}

\subsection{Architecture}

Our pretraining scheme is depicted in \Cref{fig:model_view}. We use a deep Transformer encoder \cite{vaswani2017attention}, which consists of an embedding layer followed by a stack of \textit{self-attention layers}, as implemented by BERT \cite{devlin2018bert}. To simplify our analyses, we do not add convolutional layers or 2D attention layers before the embedding layer, though other self-attentional speech models have found these helpful \cite{dong2018speech}.

Let $\mathbf{X} = [\mathbf{x}_1; \dotsb; \mathbf{x}_T]$ denote a sequence of input features. The embedding layer is applied, after which a learned matrix of position embeddings (size $d_{\text{emb}} \times T_{\text{max}}$) are added. A self-attention layer consists of two sublayers applied in sequence. The first sublayer performs \textit{multi-head self-attention}, which computes $n_{\text{hds}}$ sets of queries $\mathbf{Q}$, keys $\mathbf{K}$, and values $\mathbf{V}$ by left-multiplying inputs $\mathbf{X}$ with learned weights $\mathbf{W_Q}^{(i)}, \mathbf{W_K}^{(i)}, \mathbf{W_V}^{(i)} \in \mathbb{R}^{d_{\text{emb}}/n_{\text{hds}} \times d_{\text{emb}}}$. The output of the $i$-th attention head is:
\begin{equation}
\label{eq:sa}
    \mathbf{HdAtt}^{(i)} = \text{softmax}(\textbf{Q}^{(i)\top}\textbf{K}^{(i)}/\sqrt{d_{\text{emb}}}) \mathbf{V}^{(i)\top}
\end{equation}
where the softmax is applied row-wise. Heads are concatenated to give $d_{\text{emb}}$-dimensional features $\mathbf{MltHdAtt} = \mathbf{W_O}[\mathbf{HdAtt}^{(1)}; \dotsb; \mathbf{HdAtt}^{(n_{\text{hds}})}]^{\top}$ where $\mathbf{W_O} \in \mathbb{R}^{d_{\text{emb}} \times d_{\text{emb}}}$. The second sublayer learns a \textit{position-wise feed-forward network}, which at position $t$ computes
\begin{equation}
\label{eq:ffn}
    \mathbf{FFN}(\mathbf{x}_t) = \mathbf{W}_{2} \text{GELU}(\mathbf{W}_{1} \mathbf{x}_t + \mathbf{b}_{1})  + \mathbf{b}_{2}
\end{equation}
where $\mathbf{W_1}, \mathbf{W_2}^\top \in \mathbb{R}^{d_{\text{emb}} \times d_{\text{ff}}}$ for some inner dimension $d_{\text{ff}}$, and GELU is the non-linearity used in lieu of ReLU by BERT. Each sublayer has a residual connection and layer normalization, e.g., $\text{LN}(\mathbf{MltHdAtt}(\mathbf{X}) + \mathbf{X})$.

\subsection{Training criteria}

We train on the sum of two losses in our framework: $L_1$ reconstruction under masking, and connectionist temporal classification (CTC) \cite{graves2006ctc}. Inputs to our network are cepstral mean-normalized MFCCs of speech utterances. We stack every 3 frames to reduce the memory cost of long sequences \cite{salazar2019self}.

The first loss is our proposed bidirectional predictive coding and denoising task. At 5\% of the positions, we mask a span of 3 stacked frames (representing 9 frames or $\sim$100ms of audio). These are replaced with zero vectors (recall inputs are mean-normalized), a method proposed in the supervised setting by SpecAugment \cite{park2019specaug}. In this way we mask $\sim$15\% of tokens, similar to BERT's pretraining scheme \cite{devlin2018bert}. However, we follow SpanBERT \cite{joshi2019spanbert} in masking \textit{spans} of inputs instead, to induce higher-level representations by increasing the difficulty of the task (as speech frames are highly correlated over time). We also follow RoBERTa \cite{liu2019roberta} and generate new masking patterns for each batch. Finally, instead of reconstructing only masked positions, we reconstruct everywhere to induce acoustic information at all positions and more explicitly train {\Model} as a type of denoising autoencoder \cite{chung2016audio}:
\begin{equation}\label{eq:decoar}
  \begin{aligned}
\mathcal{L}_{\text{recons}}=\frac{1}{T} \sum_{t=1}^{T}|\mathbf{x}_{t} - \mathbf{FFN} (\mathbf{z}_t)|,
\end{aligned}
\end{equation}
where $\mathbf{z_t}$ are outputs of the Transformer encoder (our ``{\Model} vectors''). $\mathbf{FFN}$ follows Eq.~\ref{eq:ffn} but here taking ReLU, $d_{\text{ff}} = d_{\text{emb}}$, and matching $\mathbf{x}_{t}$'s dimension.

The second loss is supervised, to induce phonetic understanding and leverage our use of labeled speech corpora. To support end-to-end pretraining (i.e., no force-alignment), we propose the use of sequence-level CTC loss. Let $Y = [y_1; \dotsc; y_U]$ be the output sequence of tokens $(U \le T)$. Recall that CTC is a maximum likelihood objective $\mathcal{L}_{\text{CTC}} = - \log P(Y \mid \mathbf{X})$, where likelihood is given by summing over probabilities of all $T$-length symbol sequences $S$ (over the original label alphabet augmented with a special blank token) that collapse to $Y$ after merging repetitions and removing blanks (the operation represented by $\mathcal{B}$):
\begin{equation}
	P(Y \mid \mathbf{X}) = \sum_{S \in \mathcal{B}^{-1}(Y)} \  \prod_{t=1}^{T}\ P(s_t | \mathbf{X}).
\end{equation}
We see that position-wise tokens in the ``path'' $S$ are treated as conditionally independent from one another. The probabilities $P(s_t | \mathbf{X})$ are given by taking $\mathbf{z}_t$ and applying a linear projection then softmax.

The combined loss is
\begin{equation}
  \begin{aligned}
\mathcal{L}= \recon \cdot \sqrt{T} \cdot \mathcal{L}_{\text{recons}} + (1 - \recon) \cdot \mathcal{L}_{\text{CTC}},
\end{aligned}
\end{equation}
where $\sqrt{T}$ empirically rescales $\mathcal{L}_{\text{recons}}$ (which is averaged over frames) to be proportionate with $\mathcal{L}_{\text{CTC}}$ (a sequence-level loss), and where $\recon$ is a hyperparameter.

\section{Task-specific model}
\label{sec:task}

\begin{figure}[th!]
  \centering
  \includegraphics[trim=0 150 0 0,clip,width=1.0\linewidth]{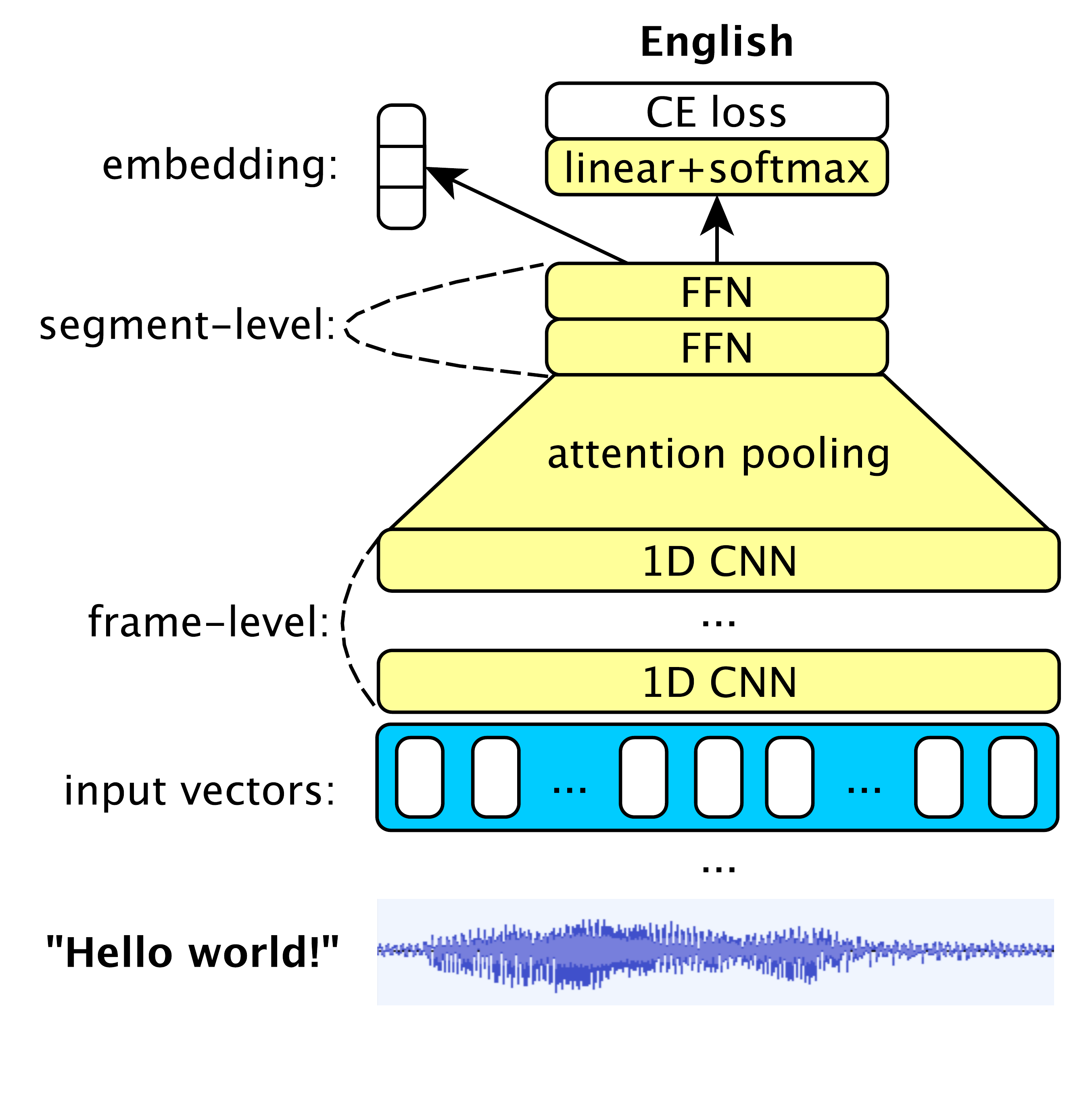}
  \caption{Architecture of our {\TaskSys}, depicted here for closed-set LR. For SR we also train with cross-entropy (CE) loss, but we extract embeddings at the final layer then use PLDA for classification.}
  \label{fig:xvector}
\end{figure}

Although our pretraining is BERT-like, our method of application is closer to ELMo \cite{peters2018deep} representations in NLP, and to bottleneck and tandem features in SR and LR: \textit{we keep {\Model} frozen}. Furthermore, we only use the final layer's output, to see whether our multitask setup has captured a balance of acoustic and phonetic information. Instead, we treat {\Model} vectors as features (similar to wav2vec \cite{schneider2019wav2vec} or DeCoAR \cite{1912.01679}) and train a segment-level DNN classifier on them.

\subsection{Architecture}

Our classifier (\Cref{fig:xvector}) is based on the DNN used to extracting utterance-level \textit{x-vectors} for speaker and language recognition \cite{snyder2018spoken, snyder2018spoken1}. We re-implement the x-vector architecture in MXNet with two substitutions: The first 5 time-delay neural network (TDNN) layers are replaced with 5 layers of 1D convolutions (CNN) with corresponding kernel sizes (2,2,3,1,1) and channel counts (512, ..., 512, 512$\times$3). Then, instead of extracting mean and standard deviation on the frame-wise vectors, we perform \textit{multi-head self-attentive pooling} (SAP) \cite{lin2017structured}. By assigning relative importance to each time step, SAP focuses on frames relevant to the utterance-level decision. SAP improves duration robustness for speaker verification \cite{liu2018duration} and is used in state-of-the-art end-to-end LR models \cite{cai2018exploring}. The output of a single head is a 512-dim.\ vector:
\begin{equation}
    \mathbf{v_{\text{SAP}}}^{(i)} = \text{softmax}(\mathbf{u}^{(i)} \tanh(\mathbf{W} \mathbf{X})) \mathbf{X}^\top.
\end{equation}
This is analogous to the self-attention layer (Eq.~\ref{eq:sa}), with $\mathbf{u}^{(i)}$ is the head's query weight vector and $\mathbf{W}$ is a shared linear transformation that gives keys. We take 5 heads and get a combined 2560-dim.\ vector, similar to the x-vector system. The rest proceeds in the same way: two dense layers of size 512, then a linear projection and softmax. Batch norm and ReLU are applied between all layers except after SAP.

\subsection{Training criterion}

We train \textit{{\TaskSys}s} on {\Model} for each of LR and SR using cross-entropy loss. For closed-set LR, we use the softmax layer directly at inference time, since the LR task has fixed language categories. For SR, we use probabilistic linear discriminant analysis (PLDA) \cite{ioffe2006probabilistic} to compare pairs of speaker embeddings.

\section{Experimental setup}
\label{sec:experiments}

\begin{table*}[th]
    \centering
    \small
    \begin{tabular}{@{} l | ccc | ccc  @{}}
        \toprule
        \multirow{2}{*}{\textbf{System}} & \multicolumn{3}{c}{\textbf{(\textit{Fisher $\rightarrow$}) Fisher}} & \multicolumn{3}{|c}{(\textbf{\textit{TED-LIUM $\rightarrow$}) VoxCeleb}} \\
        & EER & minDCF08 & minDCF10 & EER & minDCF${}_{0.01}$ & minDCF${}_{0.001}$ \\
        \midrule
        i-vector & 2.10 & 0.093 & 0.334 & 5.24 & 0.493 & 0.616 \\
        x-vector & 1.73 & 0.086 & 0.363 & 3.13 & 0.326& 0.500\\
        x-vector \textit{+ phonetic vec. + multi-tasking} \cite{liu2018speaker} & 1.39 & 0.073 & 0.308 &  -& -& -\\
        x-vector \textit{+ adv.\ loss + multi-tasking} \cite{wang2019usage} & - & - & - &  3.17 & 0.336 & -\\
        \midrule
        {\TaskSys} on MFCCs & 1.50 &  0.079& 0.316 & 3.06 & 0.322& 0.514\\
        \textit{{\TaskSys} on {\Model}} & \textbf{1.23} & \textbf{0.067} & \textbf{0.268} & \textbf{2.51} & \textbf{0.300} & \textbf{0.439}\\
        \bottomrule
    \end{tabular}
    \caption{Speaker recognition results. \textit{Italicized models} use: \textbf{Left:} pretraining on Fisher for a Fisher SR task; \textbf{Right:} pretraining on TED-LIUM for the VoxCeleb SR task.  Numbers below the midline are ours; non-pretrained i-vector and x-vector numbers on the left are from \cite{liu2018speaker}, and on the right are from the Kaldi codebase. The minDCF variants are minimum detection costs as defined in \cite{liu2018speaker, wang2019usage}.}
    \label{table:results2}
\end{table*}

\begin{table*}[th]
    \centering
    \small
    \begin{tabular}{@{} l|cc|cc|cc|cc @{}}
        \toprule
        \multirow{2}{*}{\textbf{System}} &  \multicolumn{2}{c}{\textbf{3sec}} & \multicolumn{2}{|c}{\textbf{10sec}} & \multicolumn{2}{|c}{\textbf{30sec}} & \multicolumn{2}{|c}{\textbf{Total}} \\
        & $C_\text{avg}$ & EER\% & $C_\text{avg}$ & EER\% & $C_\text{avg}$ & EER\% & $C_\text{avg}$ & EER\% \\ 
        
        \midrule
        CNN SAP \cite{cai2018exploring}  & 8.59 & 9.89    & 2.49 & 4.27    & 1.09 & 2.38     & -- & -- \\
        CNN-BLSTM SAP \cite{cai2019utterance}  & 9.22 & 9.50    & 2.54 & 3.48    & 0.97 & 1.77    & -- & --\\
        CNN-LDE \cite{cai2018insights}  & 8.25 & 7.75    & 2.61 & 2.31    & 1.13 & 0.96    & -- & --\\
        \textit{i-vector DNN \cite{richardson2015unified}}  & 19.67 & --   & 7.84 & --  & 3.31 & --   & 10.27 & -- \\
        \textit{DNN tandem features \cite{cai2018insights}} & 9.85 & 7.96   & 3.16 & 1.95    & 0.97 & 0.51 & -- & --\\
        \textit{DNN phoneme posterior features \cite{cai2018insights}} & 8.00 & 6.90   & 2.20 & 1.43    & \textbf{0.61} & \textbf{0.32}   & -- & --\\
        \midrule
        {\TaskSys} on MFCCs & 19.42 & 13.21 & 12.74 & 8.25 & 10.03 & 6.53 & 14.06 & 9.42 \\
        \textit{{\TaskSys} on {\Model}} & \textbf{6.16} & \textbf{4.63} & \textbf{2.03} & \textbf{1.21} & 1.17 & 0.79 & \textbf{3.12} & \textbf{2.19} \\
        \bottomrule
    \end{tabular}
        \caption{NIST LRE07 closed-set, general LR task results for individual models. Numbers below the midline are ours. \textit{Italicized models} use models pretrained on Fisher as a source of phonetic information.}
    \label{table:results}
\end{table*}

We use Kaldi \cite{povey2011kaldi} for data preparation. Our BERT and {\TaskSys} implementations are based on MXNet's GluonNLP toolkit \cite{gluonnlp}. We release our code and models at \url{https://github.com/awslabs/speech-representations}.

\subsection{{\Model} pretraining}

We use 40-dimensional mean-normalized MFCCs (window size of 25ms, hop length of 10ms) as inputs (before stacking every 3 frames). Our {\Model} parameters, learning schedule, and training details are consistent with the BERT base described in \cite{devlin2018bert}, with 12 self-attention layers, $d_{\text{emb}} =$ 768 hidden dimensions, $d_{\text{ff}} = 4d_{\text{emb}}$, and $n_{\text{hds}} =$ 12 heads. The main difference is while BERT trains on contiguous excerpts of text, we take our variable-length speech utterances and load them in batches of 80, spread over multiple GPUs. Our warmup is over the first 3,000 batches, to a maximum learning rate of 5e-5. We train for around 30 epochs.

Since the VoxCeleb dataset for SR is sampled at 16kHz while others are sampled at 8kHz (and to match previous configurations for phonetic pretraining), we train two rate-specific {\Model} models:
\begin{itemize}
    \item \textbf{Fisher English (8kHz).} A corpus of telephone conversations \cite{fisher}, via the \textit{train} split of Kaldi's \textit{s5} recipe.
    \item \textbf{TED-LIUM (16kHz).} A corpus of English TED talks \cite{tedlium12}, via the \textit{train} split of Kaldi's \textit{s5\_r3} recipe.
\end{itemize}
To give phoneme labels for CTC on both datasets, we use the CMUdict lexicon. We omit lexical stresses to give an alphabet of 39 non-silence classes.

\subsection{Speaker recognition (SR)}

To validate the joint usability of our representations, we take  Fisher {\Model}'s representations for the text-independent Fisher SR task described in \cite{liu2018speaker}, using the same training and evaluation sets\footnote{\scriptsize \url{https://github.com/mycrazycracy/Speaker_embedding_with\_phonetic\_information}}. This selects 172 hours of data from 5,000 speakers for training, and takes a disjoint set of 1,000 speakers for evaluation. Each person is enrolled using 10 segments (about 30sec in total) and evaluated on three 3sec segments.

To evaluate the success of {\Model} on large scale and out-of-corpus SR, we take TED-LIUM {\Model} for the text-independent VoxCeleb task \cite{voxceleb2}. We use Kaldi's \textit{voxceleb/v2} recipe. The training set includes VoxCeleb1 development set and all of VoxCeleb2. The VoxCeleb1 test set is used as the evaluation set. The number of speakers in the training set is 7,146, and the number of utterances is 2,081,192 including augmentation \cite{snyder2018spoken}.

We perform stochastic gradient descent with batch size 128, momentum 0.9, weight decay 1e-4, and a learning rate of 0.01, which is decayed by 10 when validation loss plateaus.

\subsection{Language recognition (LR)}

We take Fisher {\Model}'s representations for the closed-set general LR task of the 2007 NIST LR Evaluation (LRE07), where the objective is to identify an utterance's language from a set of 14 languages. We use the \textit{train} split of the \textit{lre07/v2} recipe, which includes LRE07's training data along with CALLFRIEND, LRE96/03/05/09, and SRE08, analogous to past work \cite{cai2018exploring,cai2018insights,richardson2015unified}. The utterances are split into 4sec segments with 50\% overlap, similar to \cite{gelly2017spoken}. Every epoch, 8,000 to 12,000 segments are randomly selected per language and distributed over batches to mitigate class imbalance. We use the same optimization as in SR, though since no validation set is available we decay the learning rate when training loss plateaus, doing this twice.

At test time, VAD was applied to reduce the length of the 6,474 closed-set test utterances. These are then split into non-overlapping segments of $\sim$4sec; however, the self-attention pooling layer (\Cref{sec:task}) occurs over frame features from all segments to give a single, utterance-level language prediction.

\section{Results\footnotemark}
\label{sec:results}

\footnotetext{In a preprint (\url{https://arxiv.org/abs/1907.00457v1}), we included results that sometimes outperformed the ones here. This was an early version of {\Model} ($\lambda=0.0$) that required various internal layers and architectures for each task. In contrast, our method here uses the output layer and the same architecture for both tasks.}

\subsection{Speaker recognition}

\Cref{table:results2} shows our model's task performance in terms of equal error rate (EER) and minimum detection cost (minDCF) versus other embedding plus phonetic information approaches. In the Fisher to Fisher case, {\Model} even improves over the shared features and multitasking approach, where the phonetic extractor is learned jointly between ASR and SR \cite{liu2018speaker}. On the large-scale, out-of-corpus VoxCeleb SR task, training on TED-LIUM {\Model} gives 18\% relative reduction in EER over training directly on MFCCs. Our model also improves on recent work that uses the same pretraining set \cite{wang2019usage} via multi-tasking and adversarial training, although their x-vector baseline is weaker.

\subsection{Language recognition}

\Cref{table:results} shows the performance of our {\TaskSys{}} trained on MFCCs and on {\Model} vectors, on the LRE07 closed-set task. Performance is reported as average detection cost $C_\text{avg}$ and equal error rate (EER\%). We get significant improvements over both end-to-end and phonetically-aware systems from the past two years. We achieve state-of-the-art on the 3sec and 10sec conditions despite having only trained on $\sim$4sec segments, which is a testament to the effectiveness of self-attention in prioritizing relevant frames. Though we underperform pretrained systems at 30sec, we still improve on all end-to-end methods.

\section{Analysis}
\label{sec:analysis}

\subsection{Reconstruction versus CTC loss}

In \Cref{table:recon} we interpolate between $\recon=0$ (CTC only) and $\recon=1$ (reconstruction only). LR and SR performance is equivalent or slightly degrades when {\Model} is only trained to reconstruct. For LR, we find that CTC-only did best; any reconstruction resulted in degradation, presumably as it degraded the quality of phonetic information encoded (though all models with $\recon < 1$ remain state-of-the-art). For SR, the model does best when some CTC loss is introduced, in line with previous work on the relevance of phonetic information to SR. As expected, using vectors from a CTC-only model actively degrades SR performance. In practice, one might balance these concerns and take, e.g., $\lambda=0.2$.

\begin{table}[th]
    \centering
    \small
    \begin{tabular}{@{} l | ccc  @{}}
        \toprule
        {\textbf{Features}} & {\textbf{PER}} & {\textbf{LR EER (3sec)} } & {\textbf{SR EER} } \\
        \midrule
        {\Model}, $\recon$ = 0.0 & 11.5 & \textbf{4.63} & 5.23\\
        {\Model}, $\recon$ = 0.2 & 13.1 & 5.19 & \textbf{1.23}\\
        {\Model}, $\recon$ = 0.5 & 13.9 & 5.42 & 1.27\\
        {\Model}, $\recon$ = 0.8 & 14.4 & 6.07 & \textbf{1.23}\\
        {\Model}, $\recon$ = 1.0 & -- & 14.97 & 1.50\\
        MFCC & -- & 13.21 & 1.50\\
        \bottomrule
    \end{tabular}
    \caption{{\Model} models pretrained on Fisher with different interpolation weights, with their phoneme error rates (PER) on the development set, plus EERs after training on downstream SR and LR tasks.}
    \label{table:recon}
\end{table}

\subsection{Intermediate {\Model} representations}
\label{ssec:downstream}

While utterance-level SR and LR are both classification problems on speech, one would expect different features to be discriminative for each task. Instead of multi-tasking to induce a balance of features in the last layer, one could instead learn to take a linear combination of features across layers. Inspired by ELMo \cite{peters2018deep}, we train the {\TaskSys{}} to instead use a global, softmax-normalized set of learned weights $s_{\ell}$ to pool representations over the $L = 12$ layers.
\begin{equation}
    \mathbf{z}_t^{\text{task}} = \sum_{\ell=1}^{L} s^{\text{task}}_{\ell} \mathbf{z}_{t, \ell}^{\text{pretrain}}.
\end{equation}
Since each $ \mathbf{z}_{t, \ell}^{\text{pretrain}}$ is layer-normalized, we interpret the weights without rescaling.

\begin{figure}[th]
  \centering
  \includegraphics[width=\linewidth]{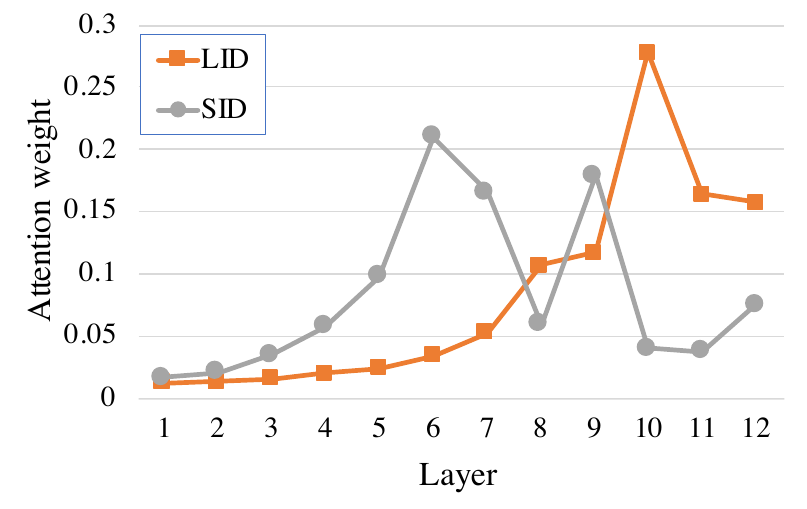}
  \caption{Per-layer weights $s^{\text{task}}_{\ell}$ learned for SR and LR tasks when trained across representations from Fisher {\Model}, $\recon$ = 0.2.}
  \label{fig:feat_attn}
\end{figure}

In \Cref{fig:feat_attn}, we see that given this flexibility, LR uses representations largely from later layers, peaking at layer 10. This is consistent with LR primarily using phonetic information. We speculate that layer 11 and 12 begin to specialize in preparation for the CTC objective (so that conditional dependence between positions are captured before the output layer). In contrast, SR uses more of the middle layers, with modes at layer 6, 9, and 12. This suggests a healthy balance of acoustic and phonetic information being leveraged. In all, this matches one's intuition that LR uses higher-level features (e.g., a language's preferred phonetic sequences) while SR uses primarily lower-level features (qualities like pitch and vocal range), plus possible phonetic preferences (given our text-independent setting).

Finally, we note that these models did not perform any better than using the last layer. We speculate that a weighted summation on fixed vectors is rather unnatural and requires an unfrozen {\Model} to perform well.

\subsection{Choice of pretraining alphabet}

We evaluate how the choice of label set for CTC affects downstream performance. We pretrain two additional models (Fisher {\Model}, $\recon$ = 0.2) with other label sets: phonemes with lexical stress (primary and secondary) using CMUdict, and characters (uncased letters, digits, space, punctuation). We train task-specific models atop these systems as before; results are in \Cref{table:unit}.
\begin{table}[th]
    \centering
    \small
    \begin{tabular}{@{} l | ccc @{}}
        \toprule
        \textbf{Pretraining alphabet} & \textbf{LR EER (3sec)} & \textbf{SR EER} \\
        \midrule
        characters & 6.44 & 1.36 \\
        phonemes (no stress) & \textbf{5.19} & \textbf{1.23} \\
        phonemes (with stress)  & 5.33 & 1.26 \\
        \bottomrule
    \end{tabular}
    \caption{Downstream results for different choices of CTC labels, when pretraining Fisher {\Model}, $\recon$ = 0.2.}
    \label{table:unit}
\end{table}

As expected, performance improves when going from characters to context-independent phonemes, the latter being more conditionally-independent (the CTC assumption) and whose prediction more explicitly encodes phonetic information. The character-based model remains competitive, which is not too surprising as character CTC is known to still learn phonetic representations internally \cite{belinkov2017analyzing}. We see equal or slight degradation when using phonemes with lexical stresses indicated.

We note that all three models have token error rates between 10\% to 13\% (upon greedy decoding via CTC), corresponding with their alphabet size. Hence, with WFST-based decoding using CMUdict and a language model, {\Model} could be easily adapted to give a competitive end-to-end ASR model that looks similar to a self-attention + CTC system (SAN-CTC) \cite{salazar2019self}.

\section{Conclusion}
We introduced {\Model}, a self-attentive, phonetically-aware, acoustic contextual representations which can be used with small task-specific models to jointly improve performance on multiple speech tasks, namely language and speaker recognition. Future work can explore the additional gains from unfreezing {\Model} as done in the original BERT work, although this removes the multi-functional property of our representations. In addition to tuning $\lambda$, one could also try using intermediate layers to improve performance.

One can also evaluate the use of {\Model} for speech recognition pretraining by adding further layers and implementing CTC decoding. Finally, note that the $L_1$ loss can be used by itself on unlabeled audio, suggesting the possibility of training on larger, unlabeled audio corpora.

\section{Acknowledgements}

We thank Davis Liang and Sundararajan Srinivasan for helpful feedback regarding this work.

\bibliographystyle{IEEEbib}
\bibliography{Odyssey2020_BibEntries}

\end{document}